\documentclass[runningheads]{llncs}
\usepackage{graphicx}
\usepackage{epsfig}
\usepackage{amsmath}
\usepackage{amssymb}
\usepackage{multirow, ctable, booktabs, makecell, wrapfig, epsfig, layouts, subcaption, lipsum}

\begin{document}
\setlength{\abovedisplayskip}{2pt}
\setlength{\belowdisplayskip}{2pt}
\title{No Token Left Behind: Efficient Vision Transformer via Dynamic Token Idling}
\titlerunning{Efficient Vision Transformer via Dynamic Token Idling}
%
\author{
Xuwei Xu\inst{1} \and
Changlin Li\inst{2} \and
Yudong Chen\inst{1} \and
Xiaojun Chang\inst{2} \and
Jiajun Liu\inst{3} \and
Sen Wang\inst{1}
}

\authorrunning{X. Xu et al.}
%
\institute{
School of Electrical Engineering and Computer Science, The University of Queensland, QLD 4066, Australia \\
\email{\{xuwei.xu, yudong.chen, sen.wang\}@uq.edu.au} 
\and
University of Technology Sydney, NSW 2007, Australia\\
\email{\{changlinli.ai, cxj273\}@gmail.com} 
\and
CSIRO, QLD 4069, Australia\\
\email{ryan.liu@data61.csiro.au}}
\maketitle              
\begin{abstract}
Vision Transformers (ViTs) have demonstrated outstanding performance in computer vision tasks, yet their high computational complexity prevents their deployment in computing resource-constrained environments. Various token pruning techniques have been introduced to alleviate the high computational burden of ViTs by dynamically dropping image tokens. However, some undesirable pruning at early stages may result in permanent loss of image information in subsequent layers, consequently hindering model performance. To address this problem, we propose IdleViT, a dynamic token-idle-based method that achieves an excellent trade-off between performance and efficiency. Specifically, in each layer, IdleViT selects a subset of the image tokens to participate in computations while keeping the rest of the tokens idle and directly passing them to this layer's output. By allowing the idle tokens to be re-selected in the following layers, IdleViT mitigates the negative impact of improper pruning in the early stages. Furthermore, inspired by the normalized graph cut, we devise a token cut loss on the attention map as regularization to improve IdleViT's token selection ability. Our method is simple yet effective and can be extended to pyramid ViTs since no token is completely dropped. Extensive experimental results on various ViT architectures have shown that IdleViT can diminish the complexity of pretrained ViTs by up to 33\% with no more than 0.2\% accuracy decrease on ImageNet, after finetuning for only 30 epochs. Notably, when the keep ratio is 0.5, IdleViT outperforms the state-of-the-art EViT on DeiT-S by 0.5\% higher accuracy and even faster inference speed. The source code is available at \url{https://github.com/Ackesnal/IdleViT}.

\end{abstract}

\section{Introduction}
Vision Transformers (ViTs) have demonstrated remarkable performance in various vision tasks, including classification \cite{dosovitskiy2020image,zhai2022scaling}, object detection \cite{liu2021swin,liu2022swin} and segmentation \cite{chen2022vision,fang2023eva}. Despite ViTs' achievements, the high computational complexity of ViTs hinders their deployments in real-world scenarios where computing resources are usually limited. As a result, there is a growing demand for efficient methods that strike a balance between performance and computational efficiency, enabling ViTs in resource-constrained environments.

Various approaches have been proposed to address the problem, such as constructing lightweight self-attention architectures \cite{mehta2021mobilevit,chen2021autoformer,chen2021mobile} and integrating efficient convolutions with the self-attention mechanism \cite{chen2021visformer,wu2021cvt,dai2021coatnet}. However, these methods often necessitate dedicated architecture design and training from scratch, which impose constraints on resource-constrained devices. Alternatively, some studies concentrate on reducing the computational complexity for pretrained ViTs while maintaining high performance. They identify the token redundancy issue in ViTs \cite{yuan2021tokens,rao2021dynamicvit} and point out that not all the image tokens contribute equally to the final prediction \cite{rao2021dynamicvit,liang2021evit}. Consequently, dynamic token pruning techniques \cite{rao2021dynamicvit,liang2021evit,fayyaz2022adaptive,meng2022adavit,xu2022evo,kong2022spvit} have been introduced to progressively eliminate those less informative tokens in a pretrained ViT without significantly compromising its performance.

\begin{figure}[t]
    \begin{subfigure}[t]{0.48\textwidth}
        \begin{subfigure}[t]{\textwidth}
            \setlength{\abovecaptionskip}{0.1cm}
            \centering
            \includegraphics[width=\linewidth]{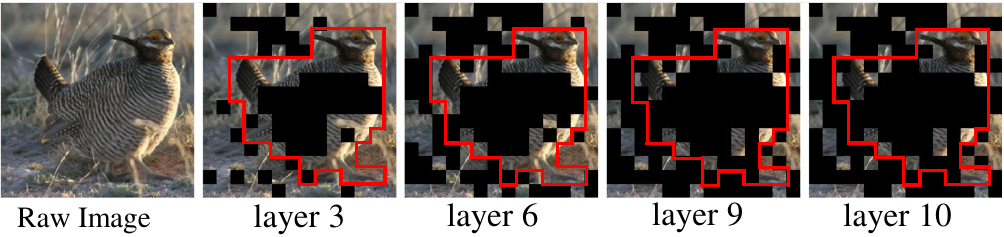}
            \vspace{-0.45cm}
            \caption{\scriptsize Token selection by DynamicViT\vspace{0.2cm}}
        \end{subfigure}
        \begin{subfigure}[t]{\textwidth}
            \setlength{\abovecaptionskip}{0.1cm}
            \centering
            \includegraphics[width=\linewidth]{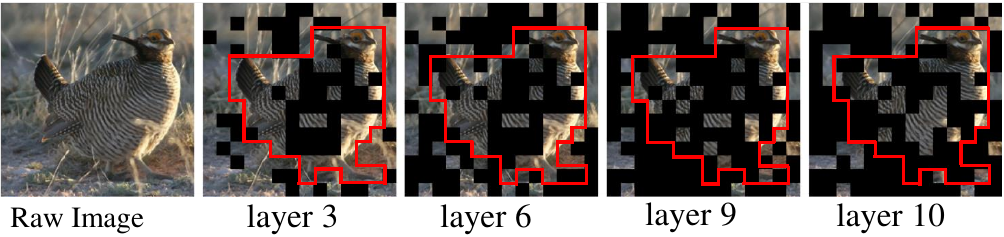}
            \vspace{-0.45cm}
            \caption{\scriptsize Token selection by IdleViT}
        \end{subfigure}
    \end{subfigure}
    \hspace{0.03\textwidth}
    \begin{subfigure}[t]{0.48\textwidth}
        \begin{subfigure}[t]{\textwidth}
            \setlength{\abovecaptionskip}{0.1cm}
            \centering
            \includegraphics[width=\linewidth]{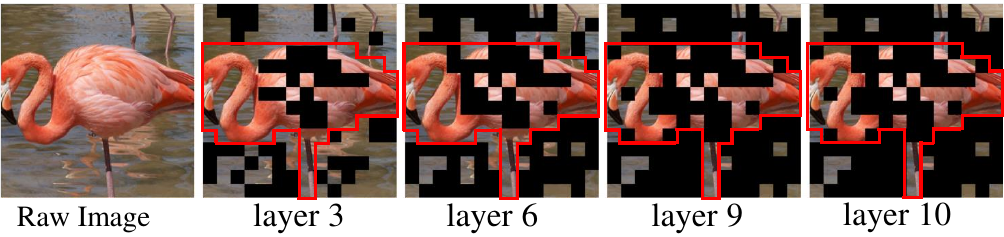}
            \vspace{-0.45cm}
            \caption{\scriptsize Token selection by DynamicViT \vspace{0.2cm}}
        \end{subfigure}
        \begin{subfigure}[t]{\textwidth}
            \setlength{\abovecaptionskip}{0.1cm}
            \centering
            \includegraphics[width=\linewidth]{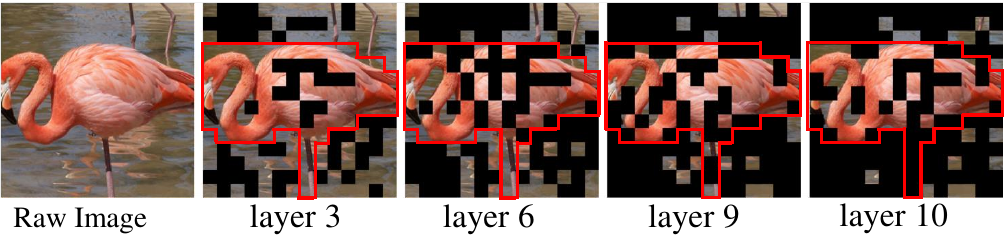}
            \vspace{-0.45cm}
            \caption{\scriptsize Token selection by IdleViT}
        \end{subfigure}
    \end{subfigure}
    \vspace{-0.3cm}
    \caption{\small\textbf{Visualized examples of self-correcting ability for IdleViT.} We take DeiT-S \cite{touvron2021training} as the backbone and compare the token selection results between a token pruning method, DynamicViT \cite{rao2021dynamicvit}, and our IdleViT. Tokens containing the foreground object have been manually labelled with red borders for comparison. IdleViT can re-select the tokens of the foreground object which are unselected in the early layers.}
  \label{fig:visualizationcompare}
\end{figure}

However, existing token pruning methods encounter an essential challenge. Empirical observations on the pruning results indicate that some tokens pruned in the early layers could be critical for accurate prediction. Unfortunately, these pruned tokens can never be re-selected in token-pruning-based methods. The information within these tokens is too early to abandon completely, yet there is no way to reintroduce them into subsequent computations. Imperfect pruning examples are illustrated in Figure \ref{fig:visualizationcompare} (a) and (c), where some important tokens of the foreground objects are pruned too early in the ViT, resulting in permanent information loss and even worse token selections in deeper layers.

\begin{figure}[t!]
    \centering
    \setlength{\abovecaptionskip}{0.05cm}
    \includegraphics[width=\textwidth, trim={0cm 0cm 0cm 0.3cm}, clip]{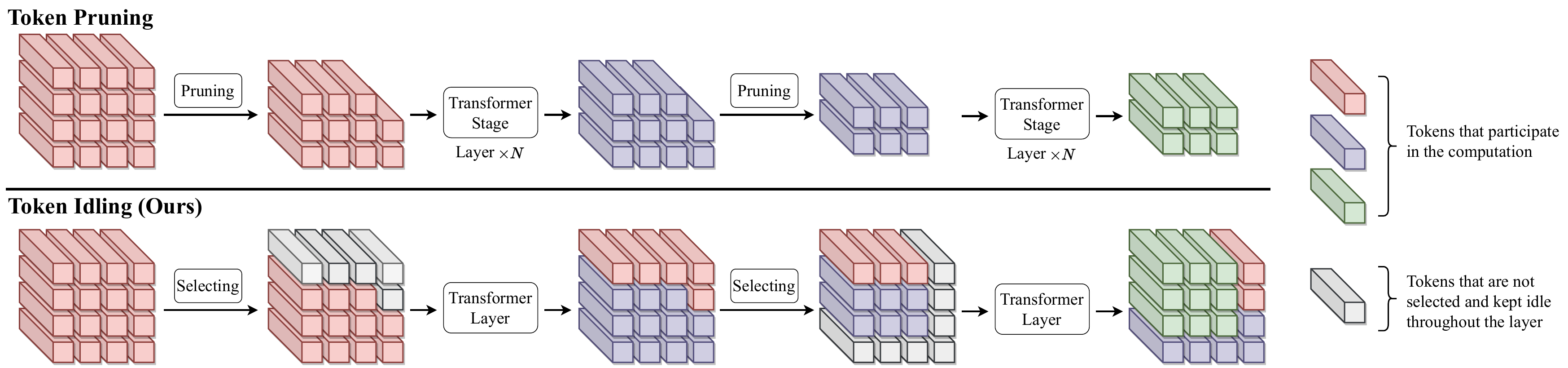}
    \caption{\textbf{Comparison between token pruning and token idling methods.} Both the token-pruning-based and token-idle-based methods reduce the number of tokens involved in computations. However, our token-idle-based method retains all the tokens, allowing the network to select previously idling tokens in subsequent layers and construct skip connections of the idling tokens.}
    \label{fig:architecturecompare}
    \vspace{-0.6em}
\end{figure}

To mitigate the aforementioned challenges, we present a novel token-idle-based efficient ViT framework, named IdleViT. Specifically, IdleViT partitions image tokens into two sets, namely the \textit{Selected} set and \textit{Idle} set, in each layer. Only the \textit{Selected} tokens participate in the self-attention computation, thereby reducing the computational complexity. The \textit{Idle} tokens remain unchanged until the end of each layer, where they are concatenated back to the \textit{Selected} tokens. Unlike previous token pruning methods, IdleViT is capable of selecting tokens from those virtually pruned in earlier layers. In Figure \ref{fig:visualizationcompare} (b) and (d), given the same keep ratio in each layer, our proposed method preserves more informative foreground patches. IdleViT also differs from the previous methods in the network's topological structure and receptive field. As Figure \ref{fig:architecturecompare} depicts, token pruning methods progressively decrease the regions visible to the ViT as the tokens are removed and eventually construct a pyramid-shaped ViT structure. On the contrary, IdleViT can completely maintain the receptive field and establish skip connections of the image tokens in the backbone. Moreover, we demonstrate that the skip connections in IdleViT can alleviate the over-smoothing problem in ViTs and contribute to the effectiveness of the token idle strategy.

Additionally, inspired by the normalized graph cut theory \cite{shi2000normalized}, we introduce a novel token cut loss as a regularization term on the attention map. The token cut loss aims to maximize pairwise attention scores within the \textit{Selected} set while minimizing attention scores between tokens from different sets. This fosters stronger intra-relationships among the \textit{Selected} tokens and restricted inter-relationships between the two sets, resulting in more distinguishable token sets. It is worth noting that the token cut loss is only applied during finetuning and does not affect the inference speed.

We have conducted extensive experiments on representative ViT models, including DeiT \cite{touvron2021training} and LV-ViT \cite{jiang2021all}, on the ImageNet-1K \cite{deng2009imagenet} dataset. The experimental results demonstrate that IdleViT can reduce ViTs' computational complexity while maintaining high accuracy. For instance, DeiT-S with IdleViT achieves 79.6\% in top-1 accuracy with a 36\% inference speed boost and a 33\% reduction in computational complexity. 

The main contributions of this paper are as follows:
\vspace{-0.5em}
\begin{itemize}
    \item We propose a novel token-idle-based efficient ViT framework called IdleViT.
    \item We devise a token cut loss as a regularization term to enhance the token partition results and improve the performance.
    \item We prove that the token idle strategy can mitigate the over-smoothing problem in existing token pruning methods.
\end{itemize}
\vspace{-0.7em}

\section{Related Work}
\vspace{-0.6em}
Vision Transformer (ViT) has attracted significant attention in the computer vision area since the success of \cite{dosovitskiy2020image}. However, the heavy computational cost of the self-attention mechanism hinders ViT's deployments in computing resource-constrained environments. As a result, dynamic token pruning methods \cite{wang2021not,rao2021dynamicvit,liang2021evit,fayyaz2022adaptive,meng2022adavit,xu2022evo,kong2022spvit} have been introduced to expedite ViTs by progressively reducing the number of tokens involved in the self-attention calculation. DVT \cite{wang2021not} achieves dynamic token numbers by early exiting from a cascade of ViTs with different token numbers. DynamicViT \cite{rao2021dynamicvit} proposes a learnable predictor to dynamically prune unimportant tokens. EViT \cite{liang2021evit} proposes a token reorganization method based on class attention without introducing extra network parameters. ATS \cite{fayyaz2022adaptive} adaptively determines the number of tokens to prune in each stage. AdaViT \cite{meng2022adavit} introduces a lightweight decision network in each block to predict whether the image patches, heads and blocks should be pruned. Evo-ViT \cite{xu2022evo} proposes a slow-fast update module that can update the tokens not involved in the computation. Different from existing dynamic token pruning methods, our IdleViT reduces the computational cost by minimizing the participation of unimportant tokens, without actually dropping them. Our method can be regarded as an extension to the previous methods.
\vspace{-0.6em}

\section{Methods}
\vspace{-0.6em}
\begin{figure}[t]
  \centering
  \setlength{\abovecaptionskip}{0.1cm}
  \includegraphics[width=0.95\textwidth]{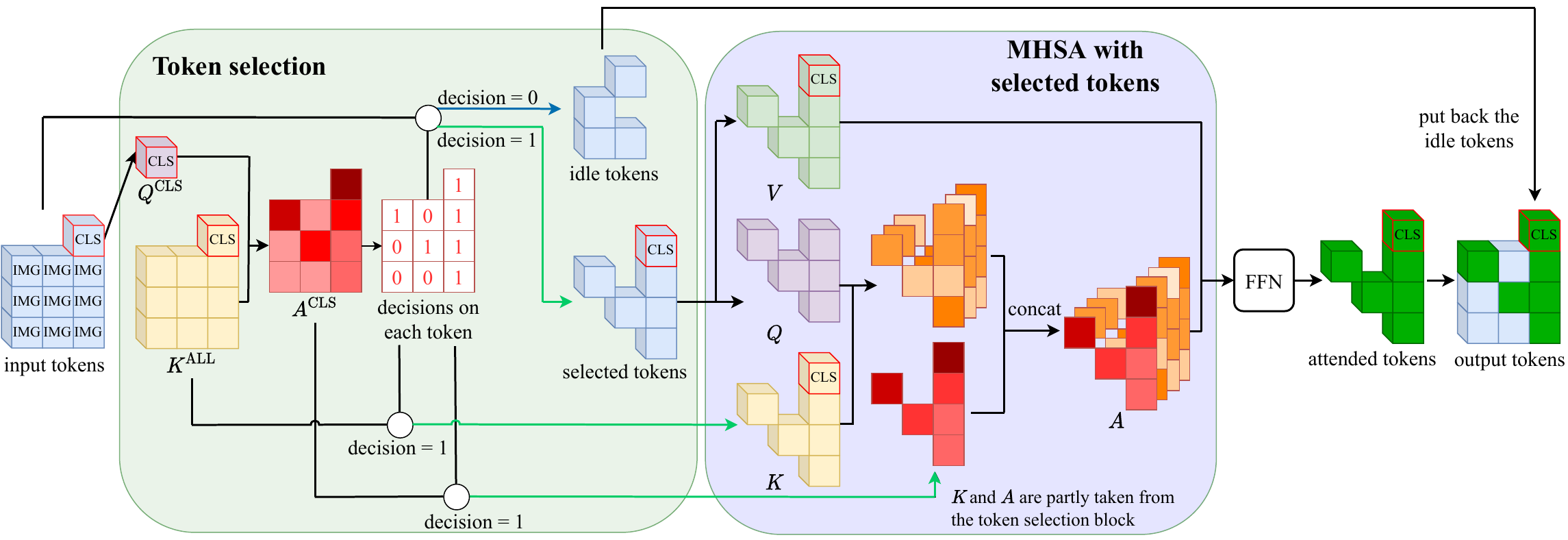}
  \caption{\textbf{IdleViT framework.} At the beginning of each layer, IdleViT selects tokens with respect to the class attention where tokens with higher attention scores are chosen to participate in computations. The idling tokens are directly passed to the output of each layer, generating the input for the next layer.}
  \label{fig:idlevit}
  \vspace{-0.5em}
\end{figure}
\subsection{Preliminaries}
\vspace{-0.3em}
\noindent \textbf{Vision Transformer (ViT).} ViT first divides and projects the input image into a number of image tokens. Analogous to the vanilla Transformer \cite{vaswani2017attention}, ViT also adds a special class token [CLS], which wraps the image information and is used for classification. Each ViT block has two layers: the multi-head self-attention (MHSA) layer and the feed-forward network (FFN) layer. Given an input feature map $X\in \mathbb{R}^{N \times C}$ with $N$ tokens and $C$ channels, the MHSA layer first linearly projects $X$ into Key ($K$), Query ($Q$) and Value ($V$) as
\begin{eqnarray}
    K = XW_{K}, \quad Q = XW_{Q}, \quad V = XW_{V},
\end{eqnarray}
where $W_{K}$, $W_{Q}$ and $W_{V}$ are the learnable weights, and the bias terms are omitted. Then, Key and Query are utilized to generate the attention ($A$) map by
\begin{eqnarray}
    \label{eq:softmax}
    A = \text{softmax}(\frac{QK^\top}{\sqrt{d_K}}),
\end{eqnarray}
where $d_K=C$ is the dimension of channels and $A\in \mathbb{R}^{N \times N}$ is usually considered as the relationships between each pair of the tokens. Finally, MHSA computes the self-attended feature map by the attention-weighted sum of $V$ and employs a linear projection to activate each token.

\noindent \textbf{Normalized graph cut.} In graph theory, a cut divides the vertices in a graph into two disjoint subgraphs. Specifically, if a directed weighted graph $\mathcal{G}=(\mathcal{V}, \mathcal{E})$, consisting of a set of vertices $\mathcal{V}$ and a set of edges $\mathcal{E}$, is partitioned into two subsets $\mathcal{S}_1$ and $\mathcal{S}_2$, where $\mathcal{S}_1 \cup \mathcal{S}_2 = \mathcal{V}$, the graph cut can be measured by
\begin{eqnarray}
    \label{eq:cut}
    \text{Cut}(\mathcal{S}_1, \mathcal{S}_2) = \sum_{i \in \mathcal{S}_1, j \in \mathcal{S}_2}{\mathcal{E}_{i,j}}.
\end{eqnarray}
The minimum cut in a graph is defined as the smallest cut among all possible cuts in the given graph. Identifying the minimum cut allows us to find meaningful partitions or boundaries in the graph, which is valuable for tasks like image segmentation, network flow analysis, and clustering. To prevent potential trivial solutions where one subset barely contains vertices, the normalized graph cut \cite{shi2000normalized} is introduced as a constrained version of graph cut, formulated as
\begin{eqnarray}
    \label{eq:ncut}
    \text{NCut}(\mathcal{S}_1, \mathcal{S}_2) = \frac{\text{Cut}(\mathcal{S}_1, \mathcal{S}_2)}{\text{Assoc}(\mathcal{S}_1)} + \frac{\text{Cut}(\mathcal{S}_2, \mathcal{S}_1)}{\text{Assoc}(\mathcal{S}_2)},
\end{eqnarray}
where $\text{Cut}(\mathcal{S}_1, \mathcal{S}_2)$ represents the graph cut between the two subsets and $\text{Assoc}(\mathcal{S}_i)$ is the association of $\mathcal{S}_i$ that ensures the scale of each subset is nontrivial. The association of subset $\mathcal{S}_i$ is defined as the sum of weights of all the edges touching vertices in $\mathcal{S}_i$, which is formulated as
\begin{eqnarray}
    \label{eq:association}
    \text{Assoc}(\mathcal{S}_i) = \sum_{j\in \mathcal{S}_i, k\in \mathcal{V}}{\mathcal{E}_{j,k}}.
\end{eqnarray}
The normalized graph cut was widely used as an optimization method for the image segmentation task. In image segmentation, methods based on the normalized graph cut usually aim to identify semantically consistent components by minimizing the normalized cut of image pixels.
\vspace{-0.7em}

\subsection{Token Selection and Idling}
\vspace{-0.3em}
As illustrated in Figure \ref{fig:idlevit}, the IdleViT framework reduces the computational cost for ViTs by dynamic token selection at the beginning of each layer and preserves the token information by token idling throughout the layer.
\vspace{-1em}

\subsubsection{Dynamic token selection.}
To determine the most informative tokens that participate in the calculation, we utilize the attention scores between the [CLS] token and image tokens by default. At the beginning of a ViT block, image tokens with top-$\mathcal{K}$ attention scores towards the [CLS] token are selected and straightforwardly named the \textit{Selected} tokens. We relax the \textit{Selected} token set to include the [CLS] token itself, which always involves in the MHSA calculation. Meanwhile, the unselected image tokens remain the same throughout the layer and are called \textit{Idle} tokens. Specifically, we denote the set of \textit{Selected} token indices and \textit{Idle} token indices as $S$ and $I$, respectively. Consequently, the \textit{Selected} tokens and \textit{Idle} tokens are denoted by $X_{S}$ and $X_{I}$, respectively.

It is worth noting that the token idle strategy is independent of any particular token selection algorithm. Instead, it can be incorporated into different token selection methods, such as the token predictor in DynamicViT \cite{rao2021dynamicvit}. By default, we apply a parameter-free method in accordance with recent research \cite{liang2021evit}, which achieves better performance and does not expand the model size.

Additionally, we use $\mathcal{K}$ to represent the number of \textit{Selected} tokens and lowercase $k$ as the base keep ratio so that $\mathcal{K} = \lfloor kN \rfloor$. Following the previous work, IdleViT also selects tokens in a hierarchical manner where the keep ratio decreases geometrically when the layer increases. We divide a ViT network into four stages and set the real keep ratio for layers in the $i^\text{th}$ stage as $k^{i-1}$.
\vspace{-1em}

\subsubsection{Token idling.} At the end of each layer, the \textit{Idle} tokens are concatenated back to the feature map based on the corresponding indices $I$. Notably, the input size and output size of a feature map in one layer remain constant.
\vspace{-0.8em}

\subsection{Token Cut Loss}
\vspace{-0.3em}
Minimizing the normalized graph cut has been recognized to enhance semantic consistency in various computer vision tasks. IdleViT separates tokens into two sets (i.e., \textit{Selected} and \textit{Idle} sets), which is analogous to a binary graph cut. Therefore, we can approximate the most semantically consistent separation of tokens by achieving the minimum normalized cut of tokens, with a slightly constrained setting where the sizes of the two sets are fixed. Based on the normalized graph cut described in Equation \ref{eq:ncut}, we propose a token cut loss on the attention map to enhance the semantic consistency of the \textit{Selected} tokens. In the scenario of ViT, the attention map $A$ can be regarded as the edge set $\mathcal{E}$ in Equation \ref{eq:cut} since they are naturally similar to each other with all non-negative values reflecting the relationships between data points.

We apply the normalized cut on the attention map with two adjustments to accommodate its peculiarities compared to the graph edges. One primary concern is the speciality of the [CLS] token. Enforcing the attention scores from the [CLS] token to \textit{Idle} tokens to be 0 can cause training collapse since the [CLS] token encodes global information. Therefore, we relax this constraint to allow the [CLS] token to have non-zero attention towards the \textit{Idle} tokens. After substituting $\mathcal{E}$ with $A$, the normalized cut on the attention map for the \textit{Selected} set $S$ and the \textit{Idle} set $I$ is formulated as
\begin{equation}
    \small
    \text{NCut}(S, I)=\frac{\text{Cut}(S, I)}{\text{Assoc}(S)} + \frac{\text{Cut}(I, S)}{\text{Assoc}(I)} =\frac{\sum_{i\in S}{\sum_{j\in I}{A_{i,j}}}}{\sum_{g\in S, h\in U}{A_{g,h}}} + \frac{\sum_{i\in I}{\sum_{j\in S\backslash \{0\}}{A_{i,j}}}}{\sum_{g\in I, h\in U}{A_{g,h}}},
    \label{eq:ncutforattention}
\end{equation}
where $U=S\cup I$ is the universal set of tokens.

Besides, we analyze the association denominator in Equation \ref{eq:ncutforattention}. Due to the softmax function in Equation \ref{eq:softmax}, the sum of attention scores towards a single token is always 1. Therefore, the two associations can be simplified as
\begin{equation}
    \small
    \label{eq:associationsimplification1}
    \text{Assoc}(S) = \sum_{g\in S, h\in U}{A_{g,h}} = \sum_{g\in S}(\sum_{h\in U}A_{g,h}) = \sum_{g\in S}1 = |S| = \mathcal{K},
\end{equation}
\begin{equation}
    \label{eq:associationsimplification2}
    \text{Assoc}(I) = \sum_{g\in I, h\in U}{A_{g,h}} = \sum_{g\in I}(\sum_{h\in U}A_{g,h}) = \sum_{g\in I}1 = |I| = N-\mathcal{K},
\end{equation}
where $\mathcal{K}$ and $N$ are the pre-defined number of selected tokens and the total number of tokens, respectively. Eventually, by taking Equation \ref{eq:associationsimplification1} and \ref{eq:associationsimplification2} into \ref{eq:ncutforattention}, we propose an inter loss to minimize the modified normalized cut as
\begin{equation}
    \begin{aligned}
        L_{\text{inter}}=\frac{1}{\mathcal{K}}\sum_{i\in S}{(\sum_{j\in I}{A_{i,j}})^2} + \frac{1}{N-\mathcal{K}}\sum_{i\in I}{(\sum_{j\in S \backslash \{0\}}{A_{i,j}})^2}.
    \end{aligned}
\end{equation}

Moreover, since our approach replaces the association term in the normalized cut with a fixed scalar, the original constraint for maximizing connections within a set no longer applies. Therefore, we introduce an additional intra loss to reinforce the intra-relationship of the \textit{Selected} set during MHSA computation. The intra loss minimizes the distance between 1 and the sum of attentions for each selected token with other selected tokens as
\begin{eqnarray}
    L_{\text{intra}}=\frac{1}{\mathcal{K}}\sum_{i\in S}{(1-\sum_{j\in S}{A_{i,j}})^2}.
\end{eqnarray}

Finally, the token cut loss for training IdleViT is the sum of intra loss and inter loss in each layer as
\begin{eqnarray}
    L_{\text{cut}} =\sum{(L_\text{intra}+L_\text{inter})}.
\end{eqnarray}
And we would like to emphasize that the token cut loss is only applied during finetuning as a regularization term and does NOT influence the inference speed. \vspace{-2em}

\subsection{Finetuning}
\vspace{-0.4em}
As stated in the Introduction section, our study aims to expedite ViTs for computing resource-constrained environments, where the training cost is also a significant constraint. Due to this scenario, our method is designed to work on pretrained ViTs with a few finetuning epochs, which is remarkably more computing resource-friendly than training from scratch. 

We follow the conventional finetuning pipeline as \cite{rao2021dynamicvit} to incorporate knowledge distillation as a training technique, where the full-size ViT model is utilized as the teacher to distil its pruned version. We employ knowledge distillation on both the logits and features, and apply the token cut loss as an additional optimization target. As a result, the total objective of finetuning IdleViT on a pretrained ViT is formulated as
\begin{eqnarray}
    L = L_{\text{cls}} + \alpha L_{\text{logit}} + \beta L_{\text{feature}} + \theta L_{\text{cut}},
\end{eqnarray}
where $L_{\text{cls}}$, $L_{\text{logit}}$, $L_{\text{feature}}$ and $L_{\text{cut}}$ are the cross entropy loss between output logits and ground truths, the KL divergence on output logits between the teacher and student model, the mean squared error on token features between the teacher and student model, and our proposed token cut loss, respectively. $\alpha$, $\beta$ and $\theta$ are the coefficients for the these loss functions.

Notably, IdleViT does not actually select a subset of tokens to participate in the calculation during finetuning. Instead, the MHSA layer calculates the full-size attention map of $X$ for the token cut loss. After each MHSA layer, we filter the \textit{Idle} tokens in the output feature map and retain them the same as the input ones. On the contrary, during testing, IdleViT only performs the MHSA and FFN on the \textit{Selected} tokens, as Figure \ref{fig:idlevit} illustrates.
\vspace{-0.6em}

\begin{table}[t]
  \caption{\textbf{IdleViT main results.} We report the accuracy, computational complexity (measured in GMACs) and inference speed (measured in image/second) of IdleViT. The blue values reflect the differences compared to the full-size model.}
  \label{tab:IdleViTmainresult}
  \centering
  \scriptsize
  \setlength{\tabcolsep}{4pt}
  \renewcommand\arraystretch{0.7}
  \begin{tabular}{lccccc}
    \toprule[1pt]
    Model & Keep ratio & Top-1 acc (\%) & Top-5 acc (\%) & GMACs &  Speed (image/s)\\
    \midrule[1pt]
    DeiT-S \cite{touvron2021training} & 1.0 & 79.8 & 94.9 & 4.6 & 2476.8 \\
    \midrule
    \multirow{8}{*}{IdleViT-DeiT-S}& 0.9 & 79.9 \textcolor{blue}{(+0.1)} & 95.0 \textcolor{blue}{(+0.1)} & 4.0 \textcolor{blue}{(-13\%)} & 2662.1 \textcolor{blue}{(+7\%)}\\
    & 0.8 & 79.9 \textcolor{blue}{(+0.1)} & 95.0 \textcolor{blue}{(+0.1)} & 3.5 \textcolor{blue}{(-24\%)} & 3031.4 \textcolor{blue}{(+22\%)}\\
    & 0.7 & 79.6 \textcolor{blue}{(-0.2)} & 94.9 \textcolor{blue}{(+0.0)} & 3.1 \textcolor{blue}{(-33\%)} & 3361.3 \textcolor{blue}{(+36\%)}\\
    & 0.6 & 79.3 \textcolor{blue}{(-0.5)} & 94.7 \textcolor{blue}{(-0.2)} & 2.7 \textcolor{blue}{(-41\%)} & 3693.0 \textcolor{blue}{(+49\%)}\\
    & 0.5 & 79.0 \textcolor{blue}{(-0.8)} & 94.5 \textcolor{blue}{(-0.4)} & 2.4 \textcolor{blue}{(-48\%)} & 4071.7 \textcolor{blue}{(+64\%)}\\
    & 0.4 & 78.4 \textcolor{blue}{(-1.4)} & 94.2 \textcolor{blue}{(-0.7)} & 2.1 \textcolor{blue}{(-54\%)} & 4362.7 \textcolor{blue}{(+76\%)}\\
    & 0.3 & 77.3 \textcolor{blue}{(-2.6)} & 93.6 \textcolor{blue}{(-1.3)} & 1.9 \textcolor{blue}{(-59\%)} & 4686.3 \textcolor{blue}{(+89\%)}\\
    \midrule[1pt]
    LV-ViT-S \cite{jiang2021all} & 1.0 & 83.3 & 96.3 & 6.6 & 700.7\\
    \midrule
    \multirow{6}{*}{IdleViT-LV-ViT-S} & 0.9 & 83.3 \textcolor{blue}{(+0.0)} & 96.3 \textcolor{blue}{(+0.0)} & 5.8 \textcolor{blue}{(-13\%)} & 752.5 \textcolor{blue}{(+7\%)}\\
    & 0.8 & 83.2 \textcolor{blue}{(-0.1)} & 96.3 \textcolor{blue}{(+0.0)} & 5.1 \textcolor{blue}{(-24\%)} & 855.3 \textcolor{blue}{(+22\%)}\\
    & 0.7 & 83.1 \textcolor{blue}{(-0.2)} & 96.3 \textcolor{blue}{(+0.0)} & 4.5 \textcolor{blue}{(-32\%)} & 937.5 \textcolor{blue}{(+34\%)}\\
    & 0.6 & 82.9 \textcolor{blue}{(-0.4)} & 96.2 \textcolor{blue}{(-0.1)} & 4.0 \textcolor{blue}{(-40\%)} & 1040.3 \textcolor{blue}{(+48\%)}\\
    & 0.5 & 82.6 \textcolor{blue}{(-0.7)} & 96.1 \textcolor{blue}{(-0.2)} & 3.6 \textcolor{blue}{(-46\%)} & 1131.4 \textcolor{blue}{(+61\%)}\\
    \bottomrule[1pt]
  \end{tabular}
  \vspace{-0.7em}
\end{table}
\section{Experiments}
\vspace{-0.5em}
\subsection{Implementation Settings}
\vspace{-0.5em}
\noindent \textbf{Dataset.} We choose ImageNet-1K \cite{deng2009imagenet} as the finetuning and testing dataset, which contains around 1.28 million images for training and 50 thousand images for validation. We compare the performance of IdleViT with other models that are also trained and finetuned on ImageNet-1K for fair comparisons.

\noindent \textbf{Backbone models.} Two representative ViTs, the DeiT \cite{touvron2021training} and LV-ViT \cite{jiang2021all}, are selected as the backbones for IdleViT. These two models are well-known in ViT families and are widely used as backbone models for token pruning methods \cite{rao2021dynamicvit,xu2022evo,liang2021evit}. Specifically, we only present the performance of DeiT-S (12 layers) and LV-ViT-S (16 layers) in this paper. Both the backbones are evenly divided into four stages with keep ratios [$1, k, k^2, k^3$] at each stage, respectively.

\noindent \textbf{Finetuning configurations.} We follow the same image augmentations and finetuning recipes in \cite{touvron2021training} for both DeiT-S and LV-ViT-S, but set the base learning rate to $2\times 10^{-5}$ and minimum learning rate to $2\times 10^{-6}$. We set the finetuning batch size to 1024 for the base keep ratios between 0.9-0.5 and 2048 for the base keep ratios between 0.4-0.3. The coefficients $\alpha$, $\beta$ and $\theta$ for the total loss are set to 5, 500 and 20, respectively. All the models are finetuned for only 30 epochs.

\noindent \textbf{Hardware.} We finetune IdleViT on 2 NVIDIA Tesla V100 GPUs and measure the speed on a single NVIDIA Tesla V100 GPU with the batch size fixed to 128. 
\vspace{-1.7em}

\subsection{Results}
\vspace{-0.7em}
\subsubsection{Main results.} Table \ref{tab:IdleViTmainresult} presents the main results of our method, demonstrating IdleViT's ability to reduce computational complexity with minimal accuracy loss. For example, IdleViT achieves even higher accuracy with a 24\% complexity reduction and 22\% speed-up on DeiT-S when $k=0.8$. Overall, our method is capable of cutting down a ViT's complexity by approximately 33\% while incurring no more than 0.2\% accuracy loss. Results on larger models are provided in the supplementary material.
\vspace{-0.8em}

\subsubsection{Comparisons with token pruning methods.} As stated in the Introduction, we target expediting ViTs in computing resource-constrained scenarios where the training cost is also a significant burden. Therefore, we only compare with dynamic token pruning methods that can be finetuned on pretrained ViT backbones for 30 epochs and exclude those methods which necessitate training from scratch for 300 epochs. As a result, we compare IdleViT with DynamicViT \cite{rao2021dynamicvit}, EViT \cite{liang2021evit}, ATS \cite{fayyaz2022adaptive} and Evo-ViT \cite{xu2022evo} on DeiT-S. Table \ref{tab:comparisons} evinces that our approach outperforms existing token pruning methods at all keep ratios. More comparisons on other keep ratios and finetuning costs are provided in the supplementary material. We also present comparisons of IdleViT with other ViTs and convolutional neural networks in the supplementary material.
\vspace{-0.8em}
\begin{table*}[t!]
  \caption{\textbf{Comparisons among token reduction methods on pretrained DeiT-S.} We compare the top-1 accuracy (Acc), computational complexity (measured in GMACs) and inference speed (measured in image/second). For different methods, we adjust their corresponding token reduction ratios to achieve similar computational complexity.}
  \label{tab:comparisons}
  \centering
  \renewcommand\arraystretch{0.9}
  \tiny
  \setlength{\tabcolsep}{1.85pt}
  \begin{tabular}{lcrclrclrclrcl}
    \toprule
    Method & \#Param & Acc & GMACs & Speed & Acc & GMACs & Speed & Acc & GMACs & Speed & Acc & GMACs & Speed\\
    \midrule
    & & \multicolumn{3}{c}{$k$=0.8} & \multicolumn{3}{c}{$k$=0.7} & \multicolumn{3}{c}{$k$=0.6} & \multicolumn{3}{c}{$k$=0.5}\\
    \cmidrule(lr){3-5}
    \cmidrule(lr){6-8}
    \cmidrule(lr){9-11}
    \cmidrule(lr){12-14}
    DyViT \cite{rao2021dynamicvit} & 22.8M & 79.6 & \textbf{3.4} & \textbf{3405.0} & 79.3 & \textbf{3.0} & \textbf{3889.6} & 78.5 & \textbf{2.5} & \textbf{4474.3} & 77.5 & \textbf{2.2} & \textbf{5147.3} \\
    EViT \cite{liang2021evit} & \textbf{22.1M} & 79.8 & 3.5 & 2285.5 & 79.5 & 3.0 & 2621.8 & 78.9 & 2.6 & 3045.1 & 78.5 & 2.3 & 3383.3 \\
    Evo-ViT \cite{xu2022evo} & 22.4M & 78.4 & 3.5 & 2292.9 & 78.2 & 3.0 & 2605.8 & 78.0 & 2.6 & 2997.7 & 77.7 & 2.4 & 3172.6 \\
    ATS \cite{fayyaz2022adaptive} & \textbf{22.1M} & 79.6 & \textbf{3.4} & 2035.7 & 79.2 & 3.1 & 2161.3 & 78.9 & 2.7 & 2228.6 & 78.2 & 2.3 & 2351.7\\
    IdleViT & \textbf{22.1M} & \textbf{79.9} & 3.5 & 3031.4 & \textbf{79.6} & 3.1 & 3361.3 & \textbf{79.3} & 2.7 & 3693.0 & \textbf{79.0} & 2.4 & 4071.7 \\
    \bottomrule
  \end{tabular}
  \vspace{-1.5em}
\end{table*}

\subsubsection{Results on pyramid ViT.} The token idle strategy, which is independent of the token selection method, can be regarded as an extension of the existing token-pruning models. To signify the superiority of the token idle strategy, we deploy IdleViT on a pyramid ViT, with DynamicViT \cite{rao2021dynamicvit} and Swin-Ti \cite{liu2021swin} as the token selection method and the backbone model, respectively. Table \ref{tab:swinresult} indicates that the token idle strategy improves the top-1 accuracy on ImageNet under various keep ratios compared to vanilla DynamicViT. Notably, EViT \cite{liang2021evit} and other token pruning methods based on the [CLS] attention cannot be employed in this scenario due to the absence of the [CLS] token in the Swin Transformer.
\vspace{-0.8em}

\noindent
\begin{minipage}[t]{\linewidth}
    \begin{minipage}[t]{0.4\linewidth}
        \centering
        \scriptsize
        \captionof{table}{\small \textbf{Token idle strategy on a pyramid ViT.} We choose pretrained Swin-Tiny \cite{liu2021swin} as the backbone and use a predictor as DynamicViT \cite{rao2021dynamicvit} to select tokens. We compare the finetuned accuracy with and without token idle.}
        \vspace{-0.8em}
        \label{tab:swinresult}
        \setlength{\tabcolsep}{2.2pt}
        \begin{tabular}{lcccc}
            \toprule
            Idle & k=1 & k=0.9 & k=0.7 & k=0.5 \\
            \midrule
            $\times$ & \multirow{2}{*}{81.2\%} & 78.9\% & 74.2\% & 65.1\% \\
            $\surd$ & & \textbf{79.9\%} & \textbf{79.6\%} & \textbf{79.5\%} \\
            \bottomrule
        \end{tabular}
    \end{minipage}
    \hspace{0.02\linewidth}
    \begin{minipage}[t]{0.56\linewidth}
        \centering
        \tiny
        \vspace{1.7em}
        \captionof{table}{\small \textbf{Effects of token cut loss.} We provide finetuning results on DeiT-S with and without token cut loss on various keep ratios.}
        \label{tab:ablationoncutloss}
        \setlength{\tabcolsep}{2.2pt}
        \begin{tabular}{ccccccccc}
            \toprule
            \multicolumn{2}{c}{Loss type} & \multicolumn{7}{c}{Top-1 acc (\%)}\\
            \midrule
             inter & intra & $k$=0.9 & $k$=0.8 & $k$=0.7 & $k$=0.6 & $k$=0.5 & $k$=0.4& $k$=0.3\\
            \midrule
             $\times$ & $\times$  & 79.8 & 79.7 & 79.5 & 79.0 & 78.5 & 78.0 & 76.7\\
             $\surd$ & $\times$ & 79.9 & 79.8 & 79.6 & 79.3 & 78.9 & 78.3 & 77.0\\
             $\times$ & $\surd$ & 79.9 & 79.8 & 79.6 & 79.3 & 78.9 & 78.2 & 77.0 \\
             $\surd$ & $\surd$ & 79.9 & 79.9 & 79.6 & 79.3 & 79.0 & 78.4 & 77.3\\
            \bottomrule
        \end{tabular}
    \end{minipage}
    \vspace{1.2em}
\end{minipage}

\vspace{-1em}
\subsection{Analysis of Token Cut Loss}
\vspace{-0.3em}
\subsubsection{Ablation study of the token cut loss.} 
Table \ref{tab:ablationoncutloss} shows the effects of our proposed token cut loss. The experiments demonstrate that the combination of both intra and inter loss yields an average accuracy improvement of 0.3\%, which represents a modest yet meaningful gain in this field. Moreover, the efficacy of the token idle strategy signifies as the keep ratio decreases. For instance, at the base keep ratio of 0.3, IdleViT achieves top-1 accuracy of 77.3\% with token cut loss, surpassing the finetuning results without token cut loss at 76.7\% by 0.6\%, which is a significant improvement in this field. It is worth noting that token cut loss is only adopted during finetuning and does not affect the inference speed.
\vspace{-1em}

\subsubsection{Effect on the attention map.} 
Furthermore, we provide insights into the impact of token cut loss on the self-attention mechanism through Figure \ref{fig:heatmap}, which illustrates the attention maps of the image tokens of Figure \ref{fig:visualizationcompare}(c). Figure \ref{fig:heatmap}(a) indicates that using both inter and intra loss regularization enables image tokens to concentrate on their respective sets during MHSA computation. Limited attentions between the $Selected$ set and the $Idle$ set indicate a clear separation of tokens and strong semantic consistency within each set. In contrast, Figure \ref{fig:heatmap}(b) shows that training solely with inter loss causes tokens to primarily focus on themselves, hindering global interactions in MHSA. Figure \ref{fig:heatmap}(c) illustrates cross-set attentions, where the $Idle$ set also interacts with the $Selected$ set, suggesting inadequate separability of the two sets from a semantic consistency perspective.
\vspace{-1.7em}

\begin{figure}[t!]
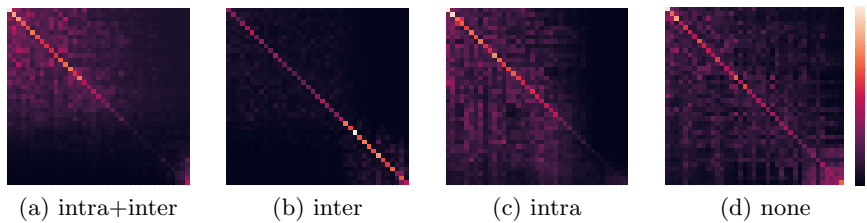

    \centering
    \begin{minipage}[b]{.2\columnwidth}
      \centering
      \centerline{\epsfig{figure=IMG/both.pdf, width=\linewidth}}
      \centerline{(a) intra+inter}
    \end{minipage}
    \hskip0.03\columnwidth
    \begin{minipage}[b]{.2\columnwidth}
      \centering
      \centerline{\epsfig{figure=IMG/inter.pdf, width=\linewidth}}
      \centerline{(b) inter}
    \end{minipage}
    \hskip0.03\columnwidth
    \begin{minipage}[b]{.2\columnwidth}
      \centering
      \centerline{\epsfig{figure=IMG/intra.pdf, width=\linewidth}}
      \centerline{(c) intra}
    \end{minipage}
    \hskip0.03\columnwidth
    \begin{minipage}[b]{.22\columnwidth}
      \centering
      \centerline{\epsfig{figure=IMG/no.pdf, width=\linewidth}}
      \centerline{(d) none}
    \end{minipage}
    \vspace{-0.5em}
    \caption{\textbf{Heat maps of the attention map finetuned with different token cut loss.} The tokens are sorted in the order of class attention score. The left-top corner represents the tokens with the highest class attention while the right-bottom corner stands for the token with the lowest class attention.}
    \label{fig:heatmap}
    \vspace{-0.8em}
\end{figure}
\begin{figure}[t!]
    \centering
    \begin{minipage}[b]{.32\columnwidth}
      \centering
      \centerline{\epsfig{figure=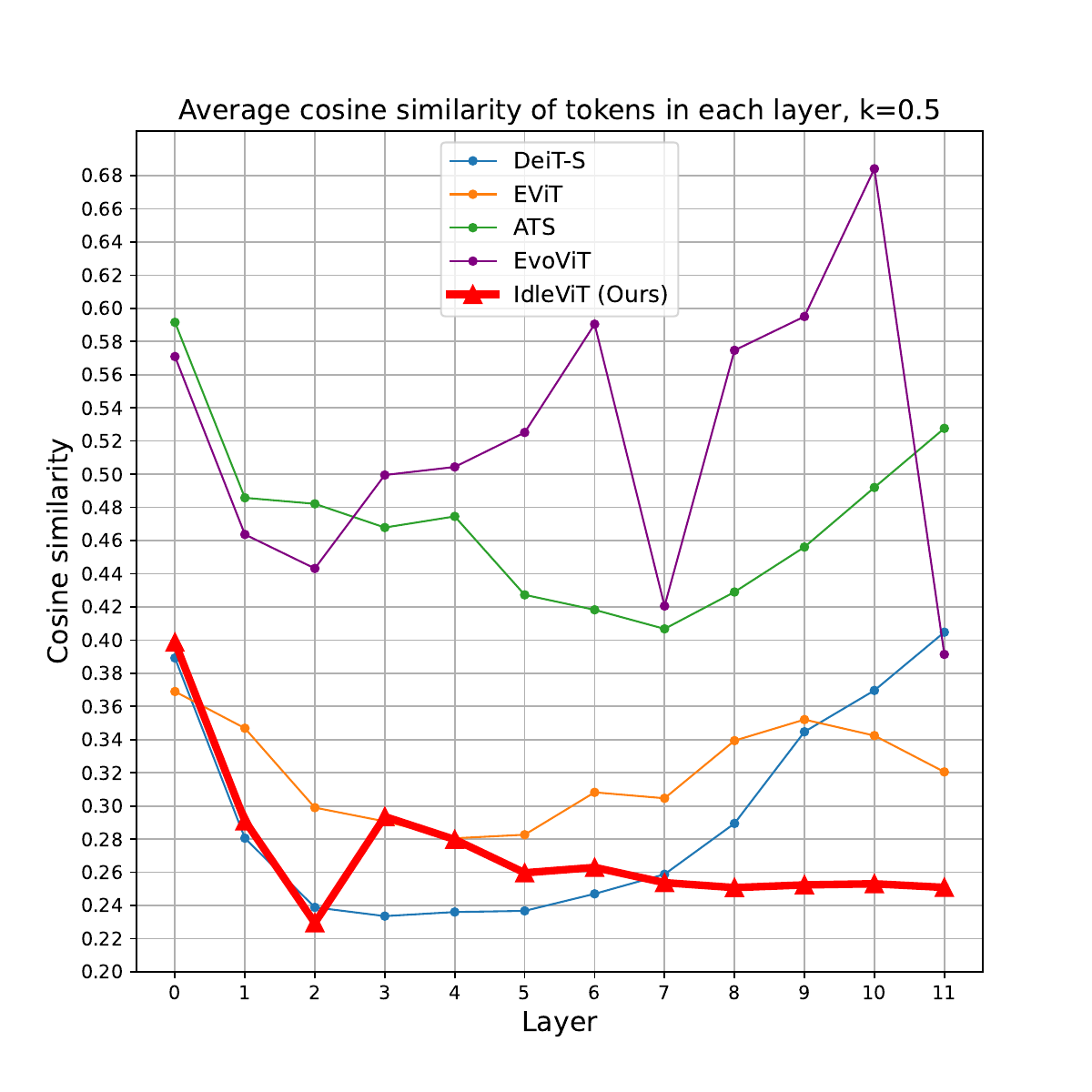, width=\linewidth}}
    \end{minipage}
    \hskip0.011\columnwidth
    \begin{minipage}[b]{.32\columnwidth}
      \centering
      \centerline{\epsfig{figure=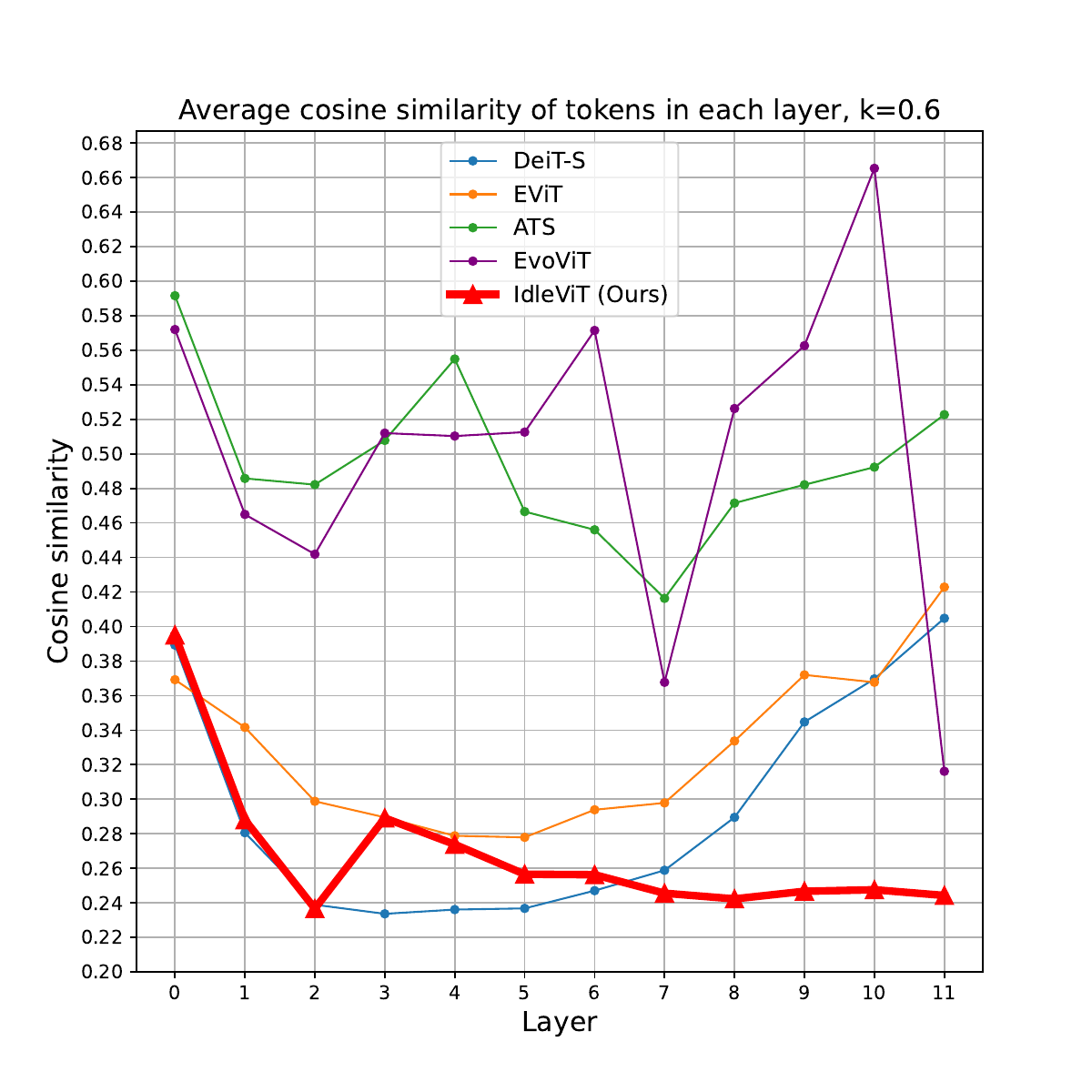, width=\linewidth}}
    \end{minipage}
    \hskip0.011\columnwidth
    \begin{minipage}[b]{.32\columnwidth}
      \centering
      \centerline{\epsfig{figure=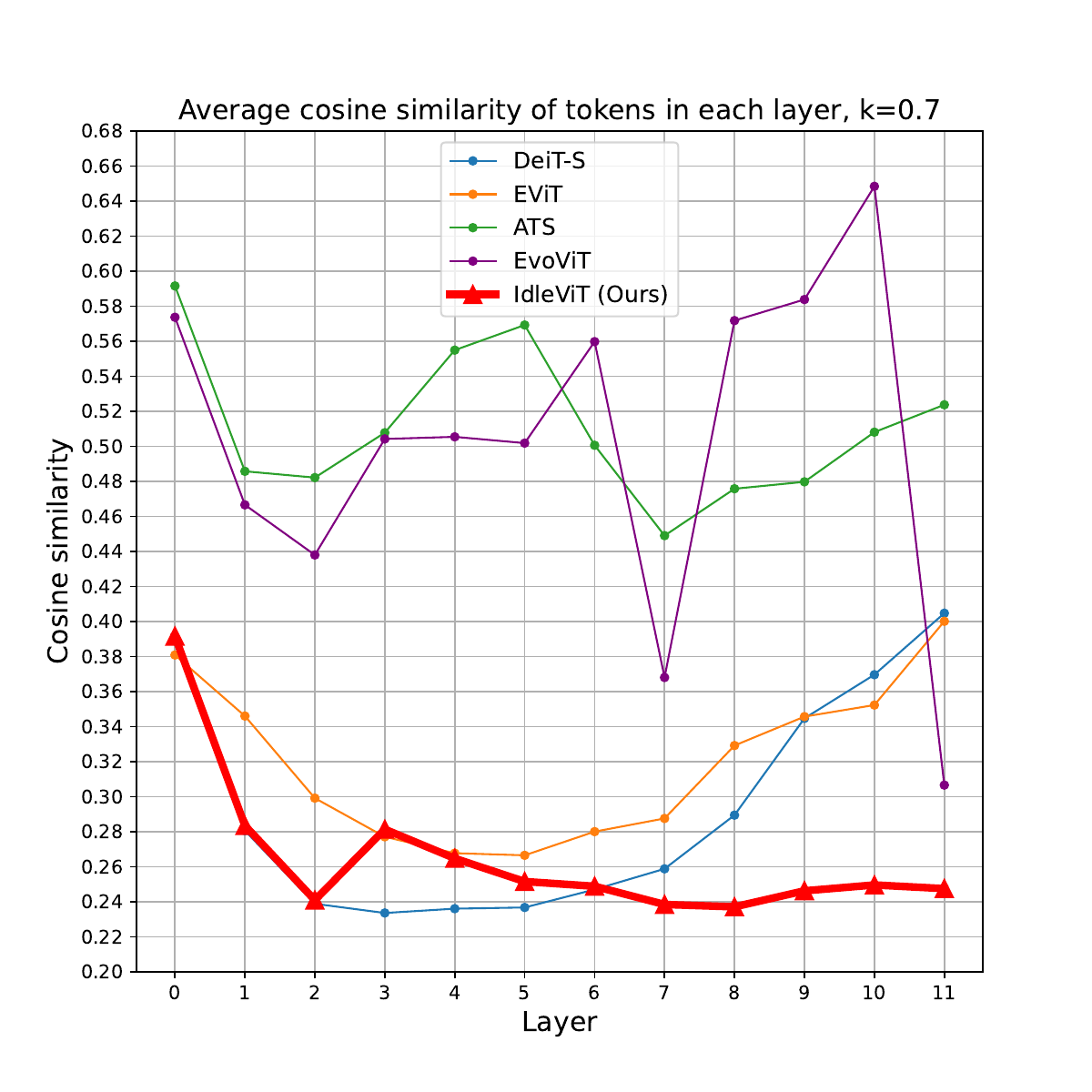, width=\linewidth}}
    \end{minipage}
    \vspace{-2em}
    \caption{\textbf{Visualization of the oversmoothing problem.} We calculate the average cosine similarity of image tokens in each layer for DeiT-S \cite{touvron2021training} (blue), EViT \cite{liang2021evit} (orange), Evo-ViT \cite{xu2022evo} (purple), ATS \cite{fayyaz2022adaptive} (green) and our IdleViT (red). A smaller average cosine similarity represents a less severe oversmoothing problem. The results clearly demonstrate that IdleViT effectively avoids increasing similarity among image tokens when compared to existing token pruning methods and vanilla DeiT-S.}
    \label{fig:heatmap}
    \vspace{-0.5em}
\end{figure}
\subsection{Analysis of Token Idle Strategy}
\vspace{-0.3em}
We investigate the reasons behind the superior performance of the token idle strategy compared to existing token pruning methods and find that IdleViT's network structure can alleviate the oversmoothing problem in ViTs. A prior study \cite{wang2022anti} observes that ViT's performance does not consistently improve with deeper layers and may even decline in very deep layers due to the oversmoothing problem. This oversmoothing problem, commonly observed in graph neural networks, results in similar image tokens as the layers deepen. In current token pruning methods, as the number of tokens progressively decreases, such oversmoothing problem becomes more severe. We compare the average cosine similarity among tokens in Table 6, where token pruning methods all lead to very similar tokens in the deep layers, which draws negative effects on the performance. However, IdleViT can reintroduce the tokens from previous layers to the deep layers and subsequently relieve the oversmoothing problem.
\vspace{-0.8em}

\section{Conclusion}
\vspace{-0.7em}
In this paper, we present IdleViT, a token-idle-based approach that reduces the computational cost of Vision Transformer without significantly compromising its performance. In each layer, a subset of tokens is selected for participation in the multi-head self-attention and feed-forward network calculation, while the unselected tokens are idled and directly sent to the end of each layer. Unlike existing token-pruning-based methods, IdleViT avoids information loss by preserving all the tokens. Additionally, we propose a token cut loss to regularize the attention map in the multi-head self-attention module, contributing to a better division of the tokens. Extensive experiments have demonstrated that our model can accelerate various ViTs with minimal accuracy loss, resulting in an excellent balance between efficiency and performance.
\vspace{-0.3em}

\bibliographystyle{splncs04}
\bibliography{main}
\end{document}


\setlength{\abovedisplayskip}{2pt}
\setlength{\belowdisplayskip}{2pt}
\title{No Token Left Behind: Efficient Vision Transformer via Dynamic Token Idling}
%
%
%
%
%
%
\appendix

\section{Implementation Details}
We conduct experiments on the ImageNet \cite{deng2009imagenet} dataset and report the test accuracy of the image classification task as the performance measurement. We finetune the pretrained model with IdleViT for 30 epochs as a standard finetuning process. When finetuning the model, we use the same configurations as DeiT \cite{touvron2021training} but set the base learning rate to 2e-5 and the minimum learning rate to 2e-6. Since we only finetune for 30 epochs, the learning rate warmup epochs are abandoned. We choose DeiT \cite{touvron2021training} and LV-ViT \cite{jiang2021all} as two backbone ViT models and embed our IdleViT framework with them in different sizes, including DeiT-S (22M parameters), LV-ViT-S (26M parameters), LV-ViT-M (56M parameters) and DeiT-B (87M parameters). In addition, the batch size is set to 1024 for DeiT-S and 512 for DeiT-B, LV-ViT-S and LV-ViT-M. We finetune the model on 2 NVIDIA Tesla V100 GPUs with 32G memory and test the speed on a single NVIDIA Tesla V100 GPU with batch size fixed to 128.

\section{Additional Experimental results}
\subsection{Effacy of IdleViT with Larger Backbones}
We conduct experiments on DeiT-B \cite{touvron2021training} and LV-ViT-M \cite{jiang2021all} with IdleViT to demonstrate the efficacy of IdleViT on larger models compared to DeiT-S and LV-ViT-S, respectively. The results are reported in Table \ref{tab:IdleViTmainresult}. IdleViT works well with large models, which can reduce the computational complexity by up to 30\% with no more than 0.3\% accuracy loss. In particular, IdleViT increases the inference speed of LV-ViT-M by 46\% with merely 0.2\% Top-1 accuracy drop.
\begin{table*}[h!]
  \caption{\textbf{IdleViT's performance on DeiT-B and LV-ViT-M.} We report the accuracy, computational complexity (measured in GMACs) and inference speed (measured in image/second) of IdleViT. The blue values reflect the differences compared to the full-size model. Due to the computing resource constraint, we only present the results for three keep ratios (0.9, 0.8 and 0.7).}
  \label{tab:IdleViTmainresult}
  \centering
  \scriptsize
  \setlength{\tabcolsep}{4pt}
  \begin{tabular}{lccccc}
    \toprule[1pt]
    Model & \makecell{Keep ratio} & Top-1 acc (\%) & top-5 acc (\%) & GMACs & Speed (image/s)\\
    \midrule[1pt]
    DeiT-B \cite{touvron2021training} & 1.0 & 81.8 & 95.6 & 17.5 & 278.8 \\
    \midrule
    \multirow{3}{*}{IdleViT-DeiT-B}& 0.9 & 81.8 \textcolor{blue}{(+0.0)} & 95.6 \textcolor{blue}{(+0.0)} & 15.2 \textcolor{blue}{(-13\%)} & 324.9 \textcolor{blue}{(+17\%)}\\
    & 0.8 & 81.5 \textcolor{blue}{(-0.3)} & 95.4 \textcolor{blue}{(-0.2)} & 13.2 \textcolor{blue}{(-25\%)} & 375.2 \textcolor{blue}{(+35\%)}\\
    & 0.7 & 81.1 \textcolor{blue}{(-0.7)} & 94.2 \textcolor{blue}{(-0.4)} & 11.5 \textcolor{blue}{(-34\%)} & 424.0 \textcolor{blue}{(+52\%)}\\
    \midrule[1pt]
    LV-ViT-M \cite{jiang2021all} & 1.0 & 84.0 & 96.7 & 12.7 & 355.3\\
    \midrule
    \multirow{3}{*}{IdleViT-LV-ViT-M} & 0.9 & 83.9 \textcolor{blue}{(-0.1)} & 96.6 \textcolor{blue}{(-0.1)} & 11.2 \textcolor{blue}{(-12\%)} & 407.4 \textcolor{blue}{(+15\%)}\\
    & 0.8 & 83.8 \textcolor{blue}{(-0.2)} & 96.6 \textcolor{blue}{(-0.1)} & 9.9 \textcolor{blue}{(-22\%)} & 465.4 \textcolor{blue}{(+31\%)}\\
    & 0.7 & 83.8 \textcolor{blue}{(-0.2)} & 96.5 \textcolor{blue}{(-0.2)} & 8.9 \textcolor{blue}{(-30\%)} & 517.1 \textcolor{blue}{(+46\%)}\\
    \bottomrule[1pt]
  \end{tabular}
\end{table*}

\subsection{Comparisons with Efficient Models}
\begin{figure}[t!]
  \centering
  \setlength{\abovecaptionskip}{0.cm}
  \setlength{\belowcaptionskip}{0.cm}
  \includegraphics[width=\columnwidth, trim={5cm 0.6cm 5cm 2.7cm}, clip]{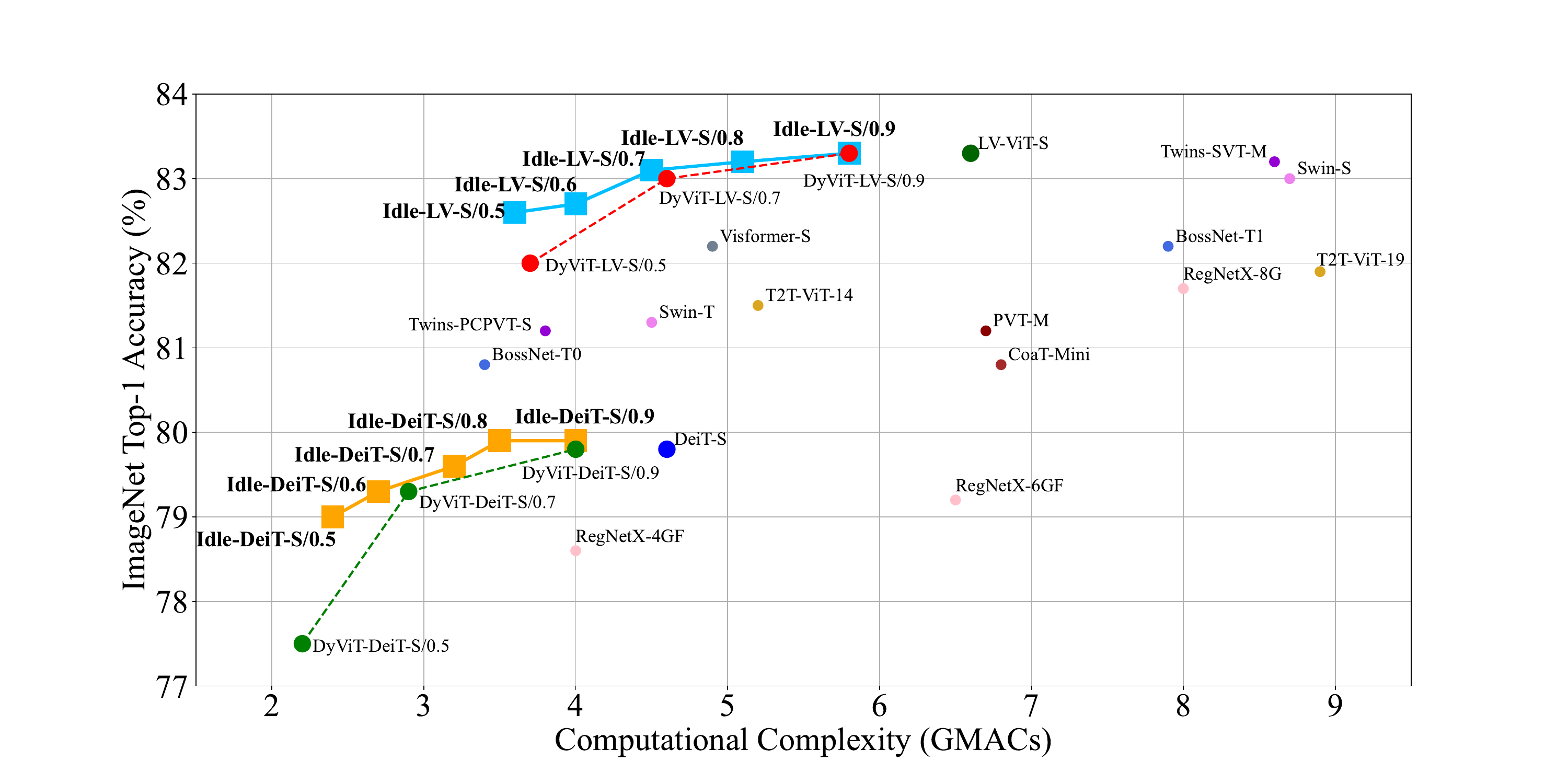}
  \caption{\textbf{Comparisons of different models on the trade-off between computational complexity and Top-1 accuracy on ImageNet.} IdleViT (squares) clearly achieves better trade-offs between complexity and accuracy than other models (circles).}
  \label{fig:comparisons}
  \vspace{-1em}
\end{figure}

Figure \ref{fig:comparisons} shows the comparisons on the trade-off between performance and computational complexity among IdleViT and other efficient ViTs and convolutional neural networks, including DynamicViT\cite{rao2021dynamicvit}, DeiT\cite{touvron2021training}, LV-ViT\cite{jiang2021all}, Swin\cite{liu2021swin}, Visformer\cite{chen2021visformer}, CoaT\cite{xu2021co}, T2T-ViT\cite{yuan2021tokens}, PVT\cite{wang2021pyramid}, Twins\cite{chu2021twins}, RegNet\cite{radosavovic2020designing} and BossNet\cite{li2021bossnas}. IdleViT clearly outperforms these efficient methods on the trade-off between performance and efficiency.

\subsection{More Comparisons with Token Pruning Methods}
We provide additional comparisons with token pruning methods when the keep ratio is set to 0.9, 0.4 and 0.3 in Table \ref{tab:comparisons}.
\begin{table*}[t!]
  \caption{\textbf{Additional comparisons among token reduction methods on pretrained DeiT-S.} We compare the top-1 accuracy (Acc), computational complexity (measured in GMACs), inference speed (measured in image/second) and finetuning cost (measured in GPU$\cdot$days). For different methods, we adjust their corresponding token reduction ratios to achieve similar computational complexity.}
  \vspace{-0.1em}
  \label{tab:comparisons}
  \centering
  \renewcommand\arraystretch{0.9}
  \tiny
  \setlength{\tabcolsep}{1.85pt}
  \begin{tabular}{lccrclrclrclrcl}
    \toprule
    Method & \#Param & \makecell{Fintuning\\cost} & Acc & GMACs & Speed & Acc & GMACs & Speed & Acc & GMACs & Speed\\
    \midrule
    & & & \multicolumn{3}{c}{$k$=0.9} & \multicolumn{3}{c}{$k$=0.4} & \multicolumn{3}{c}{$k$=0.3}\\
    \cmidrule(lr){4-6}
    \cmidrule(lr){7-9}
    \cmidrule(lr){10-12}
    \cmidrule(lr){13-15}
    DyViT \cite{rao2021dynamicvit} & 22.8M & 38.0 & 79.8 & \textbf{4.0} & \textbf{2888.8} & 76.0 & \textbf{1.9} & \textbf{5741.2} & 73.8 & \textbf{1.7} & \textbf{6420.1} \\
    EViT \cite{liang2021evit} & \textbf{22.1M} & \textbf{32.5} & 79.8 & 4.0 & 1986.3 & 77.6 & 2.0 & 3716.7 & 76.1 & 1.8 & 4106.3\\
    Evo-ViT \cite{xu2022evo} & 22.4M & 173.0 & - & - & - & 77.5 & 2.1 & 3548.0 & 76.9 & 1.9 & 3978.2\\
    ATS \cite{fayyaz2022adaptive} & \textbf{22.1M} & 49.2 & 79.8 & 4.1 & 1765.6 & 76.4 & 2.0 & 2579.8 & 74.5 & 1.8 & 2765.1\\
    IdleViT & \textbf{22.1M} & 33.0 & \textbf{79.9} & \textbf{4.0} & 2662.1 & \textbf{78.4} & 2.1 & 4362.7 & \textbf{77.3} & 1.9 & 4686.3\\
    \bottomrule
  \end{tabular}
  \vspace{-1.5em}
\end{table*}

\subsection{Ablation of Knowledge Distillation}
The effect of knowledge distillation is provided in Table \ref{tab:ablationonkd}. The results indicate that knowledge distillation can boost the finetuning of IdleViT. We also find that the improvement from knowledge distillation becomes more significant when the keep ratio goes lower.
\begin{table}[h!]
    \caption{\textbf{Results of finetuning IdleViT on DeiT-S with and without knowledge distillation.}}
    \label{tab:ablationonkd}
    \centering
    \renewcommand\arraystretch{0.8}
    \scriptsize
    \setlength{\tabcolsep}{4pt}
    \begin{tabular}{ccccc}
        \toprule
        Keep ratio & Top-1 acc (\%) & Top-5 acc (\%) & Top-1 acc (\%) & Top-5 acc (\%) \\
        \midrule
         & \multicolumn{2}{c}{with KD} & \multicolumn{2}{c}{without KD} \\
        \cmidrule(r){2-3}
        \cmidrule(r){4-5}
        0.9 & 79.9 & 95.0 & 79.7 & 94.8 \\
        0.8 & 79.9 & 95.0 & 79.6 & 94.8 \\
        0.7 & 79.6 & 94.9 & 79.3 & 94.6 \\
        0.6 & 79.3 & 94.7 & 79.0 & 94.5 \\
        0.5 & 79.0 & 94.5 & 78.6 & 94.3 \\
        \bottomrule
    \end{tabular}
\end{table}

\section{Quantitative Analysis of Token Idle Strategy}
We provide statistical results on the ImageNet validation set in Table \ref{tab:quantitativeanalysis} to substantiate IdleViT's self-correcting capability. We introduce the term "re-selected tokens" as the ones that are not selected in several intermediate layers but re-selected by IdleViT in one or more subsequent layers. 

Firstly, to study the effectiveness of token idling, we define $P(A)$ as the average percentage of re-selected tokens among all the image tokens. Table \ref{tab:quantitativeanalysis} shows that our model is capable of re-selecting tokens. For example, when keep ratio is 0.7, there are around 22\% tokens are re-selected. And the number of re-selected tokens is negatively correlated to the keep ratio.

Secondly, as the tokens involved in the last layer's computation are highly relevant to the output class token, we further analyze the re-selection rate in the last layer. Specifically, we define $P(R)$ as the average percentage of re-selected tokens tokens in the last layer over all image tokens, and $P(L)=k^3$ as the percentage of tokens in the last layer among all image tokens. $R(L)=\frac{P(R)}{P(L)}$ is the ratio of re-selected tokens to all the selected tokens in the last layer. Table \ref{tab:quantitativeanalysis} presents that around 31$\sim$36\% of the tokens involved in the last layer’s computation are somehow re-selected by IdleViT. This ratio is consistent among all the keep ratios.
\begin{table}[h!]
    \centering
    \caption{\textbf{Statistics of token recovery.} We report the token recovery statistics on ImageNet validation set. $P(A)$, $P(R)$ and $P(L)$ are respectively the percentages of recovered tokens, recovered tokens in the last layer and selected tokens in the last layer, over all the image tokens. The ratio of recovered tokens to all the selected tokens in the last layer $R(L)$ is calculated by $P(R)/P(L)$.}
    \label{tab:quantitativeanalysis}
    \scriptsize
    \renewcommand\arraystretch{0.8}
    \setlength{\tabcolsep}{6pt}
    \begin{tabular}{ccccc}
        \toprule
        Keep ratio & $P(A)$ & $P(R)$ & $P(L)$  & $R(L)$ \\
        \midrule
         0.9 & 35.3 \% & 22.6 \% & 72.9 \% & 31.0 \% \\
         0.8 & 28.1 \% & 15.9 \% & 51.2 \% & 31.1 \% \\
         0.7 & 22.1 \% & 11.3 \% & 34.3 \% & 32.9 \% \\
         0.6 & 18.1 \% & 7.8 \% & 21.6 \% & 36.1 \% \\
         0.5 & 6.9 \% & 4.1 \% & 12.5 \% & 32.8 \% \\
        \bottomrule
    \end{tabular}
\end{table}

\vspace{3em}
\section{More Visualization}
We provide more visualized comparisons between the token-pruning-based method and our IdleViT. We take DynamicViT \cite{rao2021dynamicvit} as the representative for the token-pruning-based method and use DeiT-S as the backbone. The base keep ratio is set to 0.7 for both methods. The images are picked from the ImageNet validation set. The visualizations indicate that our method is capable of recovering the informative tokens on the foreground object that are abandoned in the early stages, which lead to a more accurate classification result. The token idling approach also enables the backbone network to flexibly focus on different parts of the image in different layers.
\begin{figure}[h!]
  \centering
  \includegraphics[width=\textwidth, trim={0 0 0 0.5mm}, clip]{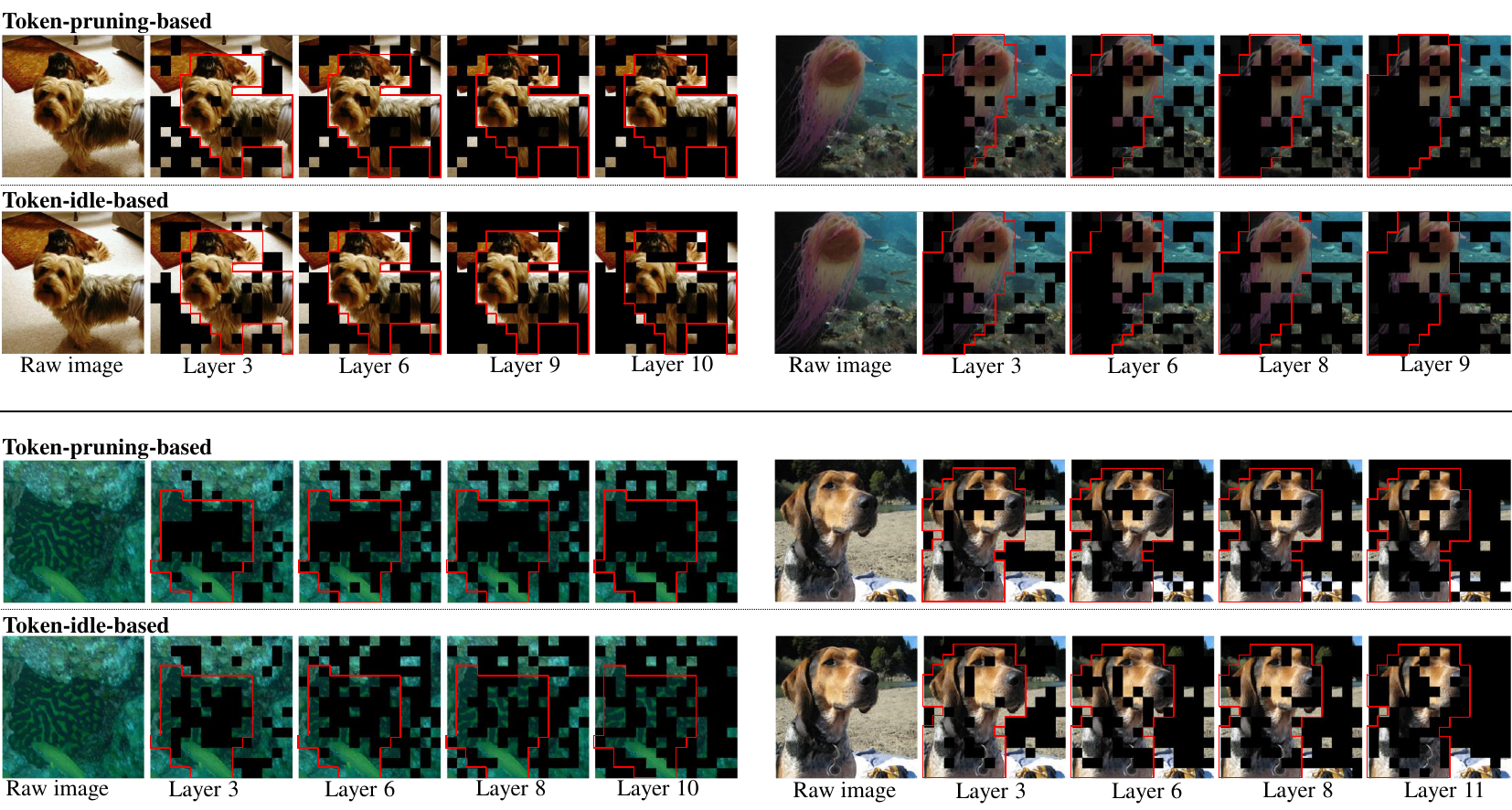}
  \caption{\textbf{Visualized examples of self-correcting ability for IdleViT.} We take DeiT-S \cite{touvron2021training} as the backbone and train it with the DynamicViT \cite{rao2021dynamicvit} and IdleViT separately. The main body of the foreground object is manually labeled for comparisons.}
  \label{fig:architecturecompare}
\end{figure}

\bibliographystyle{splncs04}
\bibliography{main}